\documentclass[letterpaper]{article} %
\usepackage[]{aaai23}  %
\usepackage{times}  %
\usepackage{helvet}  %
\usepackage{courier}  %
\usepackage[hyphens]{url}  %
\usepackage{graphicx} %
\usepackage{natbib}  %
\usepackage{caption} %
\usepackage{algorithm}
\usepackage{algorithmic}
\usepackage{microtype}

\usepackage{newfloat}
\usepackage{listings}
\DeclareCaptionStyle{ruled}{labelfont=normalfont,labelsep=colon,strut=off} %
\floatstyle{ruled}
\newfloat{listing}{tb}{lst}{}
\floatname{listing}{Listing}

\usepackage{xspace}
\usepackage{todonotes}
\usepackage{amsmath}
\usepackage{caption}
\usepackage{subcaption}
\usepackage{booktabs}
\usepackage{multirow}
\usepackage{tabularx}

\newcommand{\ie}{i.e.,\xspace}
\newcommand{\eg}{e.g.,\xspace}
\newcommand{\eat}[1]{}
\newcommand{\paratitle}[1]{\noindent\textbf{#1}\ }

\usepackage{amssymb}%
\usepackage{pifont}%

\title{A Survey on Model Compression and Acceleration \\ for Pretrained Language Models}
\author{
	Canwen Xu, Julian McAuley
}
\affiliations{
    University of California, San Diego\\

    \{cxu, jmcauley\}@ucsd.edu
}

\usepackage{bibentry}

\begin{document}

\maketitle

\begin{abstract}
Despite achieving state-of-the-art performance on many NLP tasks, the high energy cost and long inference delay prevent Transformer-based pretrained language models (PLMs) from 
seeing broader adoption
including for edge and mobile computing. %
Efficient NLP research aims to comprehensively consider computation, time and carbon emission for the entire life-cycle of NLP, including data preparation, model training and inference. In this survey, we focus on the inference stage and review the current state of model compression and acceleration for pretrained language models, including benchmarks, metrics and methodology. %
\end{abstract}

\section{Introduction}
The recent success of applying pretrained deep Transformers~\citep{transformer} on different NLP tasks~\citep{bert,t5,scao2022bloom} 
has raised concerns about its efficiency.
The high computational cost also prevents these pretrained language models (PLMs) from being deployed in production~\citep{mobilebert}.
To address this problem, \textit{efficient inference} refers to techniques that aim to make inference of an ML model faster (time-efficient), consume fewer computational resources (computation-efficient), less memory (memory-efficient) and less disk space (storage-efficient).
One popular class of techniques is \textit{model compression and acceleration}, where a large and slow model is compressed to a lightweight model that can be stored with limited disk space on a mobile device, or accelerated to run
with low latency (or both). Also, training a large model and then compressing it to a small one can be efficient for training and good for generalization~\citep{li2020train}. 

In addition to technical 
considerations,
large models also raise environmental and ethical concerns~\citep{parrot}. Large models have a  high carbon %
footprint
which
a compressed model can reduce, potentially with
little sacrifice in performance. Meanwhile, large models set obstacles for engineers and researchers from developing countries who cannot afford the necessary hardware for running the model~\citep{parrot}. Thus, model compression and acceleration can be critical to make state-of-the-art NLP techniques more accessible and facilitate inclusiveness.

\paratitle{What's covered?} In this survey, we aim to highlight the most important works in the field of model compression and acceleration for PLMs. We review the metrics, benchmarks, and methods, organizing these works in a new taxonomy. Widely-used techniques, including weight sharing, low-rank factorization, pruning, quantization, knowledge distillation, early exit, and token skipping, are covered with comparative analysis. We also highlight current challenges and future research directions in the field, calling for community efforts to build an environmentally-friendly, inclusive and sustainable future of NLP.

\paratitle{What's not covered?} This survey does not cover (1) methods that design a new architecture for training from scratch (\eg long-range Transformers, Mixture-of-Experts models); (2) data-efficient or parameter-efficient model tuning that focuses more on the training efficiency rather than inference efficiency (\eg few-shot learning, prompt learning, partial model tuning); (3) works that use the techniques surveyed in this paper but for other purposes or are application-specific (\eg self-distillation, representation distillation for retrieval).

There have been surveys~\citep{qiu2020pre,han2021pre,xu2021survey} that cover some aspects of this topic. Different from these works, we focus on the latest progress on model compression and acceleration for pretrained language models, highlighting the intersection between language technology and efficient ML.

\section{Metrics and Benchmarks}

\subsection{Metrics}
\label{sec:metrics}

There are various metrics to depict inference efficiency in different dimensions. These metrics are often reported together with accuracy to evaluate an NLP model.

\paratitle{Floating point operations (FLOPs)} directly measures the number of floating points operations needed for executing an instance. FLOPs can serve as a metric for computational efficiency and is somewhat hardware-agnostic. However, FLOPs cannot accurately reflect the real runtime since the degree of parallelism (DOP) varies for different algorithms.

\paratitle{Inference time} (\ie delay) is used to measure the runtime of an algorithm in its inference stage. Inference time can vary on different infrastructures. When testing on the same architecture, compared to FLOPs, inference time can better approximate the real-world performance of a system by taking parallelism into consideration.

\paratitle{Speed-up Ratio}
is the ratio of inference time of the baseline model to the accelerated model. Compared to inference time, speed-up ratio draws a relative comparison which can be roughly compared across different hardware. In some works~\citep{pabee,sun2021early}, speed-up ratio is approximated by the ratio of the number of Transformer layers in the baseline model to those used in calculation of an acceleration method.

\paratitle{Number of Parameters and Model Size} are often reported in NLP studies as metrics that directly reflect the storage cost of a model. This can be important for mobile deployment of an NLP model due to limited storage on a mobile device. It can also be an indicator for the memory footprint and computational cost for training and inference. An exception is models with weight sharing. For example, the FLOPs and memory use of ALBERT~\citep{albert} is slightly higher than a BERT model~\citep{bert} with the same number of layers. However, since all Transformer layers in ALBERT share the same weights, the model size of $n$-layer ALBERT is only $1/n$ of $n$-layer BERT. 

\paratitle{Carbon Footprint} measures the environmental impact. \citet{lacoste2019quantifying} provide a calculator for CO$_2$ by querying a database of carbon emission of mainstream cloud computing providers. Alternatively, Experiment Impact Tracker~\citep{henderson2020towards} and CodeCarbon\footnote{\url{https://codecarbon.io/}} are two plugins that can record the energy use of hardware and estimate the carbon emission based on the geolocation.

\paratitle{Loyalty/Fidelity} Recent works \citet{xu2021beyond} and \citet{stanton2021does} propose \textit{loyalty} and \textit{fidelity}, respectively. Both are similarity metrics calculated between the predicted distributions of the teacher and the student, in a teacher-student distillation or compression setting. \textit{Loyalty} and \textit{fidelity} can reflect how successful the knowledge transfer is from the teacher to the student, providing interpretability and reliability~\citep{xu2021beyond,stanton2021does}, and can be an indicator of better generalization in distilling large teacher models and ensembles~\citep{stanton2021does}.

\paratitle{Robustness}
\citet{su2018is} find that smaller neural networks are more vulnerable to adversarial attacks. \citet{xu2021beyond} suggest reporting adversarial robustness in addition to accuracy. In addition to adversarial robustness, \citet{du2021compressed} find compressed pretrained language models are significantly less robust on out-of-distribution (OOD) data.

\subsection{Benchmarks}
\label{sec:benchmarks}
\paratitle{Standard Benchmarks}
Most studies evaluate on common NLP benchmarks. For example, GLUE~\citep{glue} and SuperGLUE~\citep{superglue} are used for natural language understanding (NLU). SQuAD~\citep{squad} is used for machine reading comprehension (MRC).

\paratitle{EfficientQA}
EfficientQA~\citep{efficientqa} is an open-domain question answering benchmark encouraging solutions that efficiently store and access knowledge with the smallest number of bytes. EfficientQA has three resource-restricted tracks, including two tracks with a 6 GB and 500 MB cut-off for system size and one track that ranks the systems that achieves 25\% accuracy with the smallest size.

\paratitle{SustaiNLP}
The shared task of SustaiNLP 2020~\citep{wang2020overview} uses SuperGLUE~\citep{superglue} to evaluate the performance of submissions. There are three tracks that target different accuracy levels and hardware (2 GPU tracks and 1 CPU track). Within each track, submissions are ranked by lowest energy consumption, measured by Experiment Impact Tracker~\citep{henderson2020towards}.

\paratitle{ELUE}
Efficient Language Understanding Evaluation~\citep{liu2021towards} is proposed as an attempt to clearly depict the Pareto Front of FLOPs versus performance. ELUE consists of six datasets of three tasks (sentiment analysis, natural language inference, and paraphrasing). ELUE has four tracks with a parameter number cut-off of 40M, 55M, 70M and 110M. The metric used for evaluation is ELEU score, which calculates an average performance advantage over a baseline (ElasticBERT) under different FLOPs.

\begin{table*}[t]
    \centering
    \resizebox{1.\linewidth}{!}{
    \begin{tabular}{lcccc}
        \toprule
          & Hessian-based pruning & Magnitude pruning & $L_0$ regularization & Movement pruning \\
          & \citep{obd} & \citep{han2016deep} & \citep{l0} & \citep{sanh2020movement} \\
        \midrule
        Pruning Decision & 2nd order & 0th order & 1st order & 1st order \\
        Learning Objective & $\mathcal{L}$ & $\mathcal{L}$ & $\mathcal{L} + \lambda_{l0} \mathbb{E}(L_0)$ & $\mathcal{L}$\\
        Scores $\mathbf{S}$ 
        	& $-\sum_{t} (\frac{\partial^2 {\cal L}}{\partial {W^2}_{i,j}})^{(t)} {W}^{2(t)}_{i,j}$ 
            & $|W_{i,j}|$ 
            & $-\sum_{t} (\frac{\partial {\cal L}}{\partial {W}_{i,j}})^{(t)} {W}^{(t)}_{i,j} f(\overline{S}_{i,j}^{(t)})$
            & $-\sum_{t} (\frac{\partial {\cal L}}{\partial {W}_{i,j}})^{(t)} {W}^{(t)}_{i,j}$ 
            \\
        \bottomrule
    \end{tabular}
    }
    \caption{A summary of various pruning methods. $\mathbf{S}$ are saliency scores used to determine which weights to prune. The table style is borrowed from \citet{sanh2020movement}.}
    \label{tab:prune}
\end{table*}

\section{Methods}

\subsection{Weight Sharing}
Weight sharing is based on the assumption that large-scale models, like Transformer~\citep{transformer}, are over-parameterized~\citep{li2020train}. Weight sharing provides a way to decouple computation and parameters by reusing the same parameters for multiple computations. Weight sharing can reduce inference memory footprint and number of parameters and thus is memory- and storage-efficient.

\paratitle{Encoder-Decoder Sharing} In the vanilla Transformer model~\citep{transformer} for neural machine translation (NMT), there is one encoder for encoding the input into hidden representations, and one decoder for decoding it to the target language. Tied Transformer~\citep{xia2019tied} shares the weights of the encoder and decoder of Transformer. The results of Tied Transformer are comparable to the vanilla Transformer. \citet{rothe2020leveraging} leverage pretrained language model checkpoints to initialize a sequence-to-sequence model. They experiment with a shared encoder and decoder to reduce memory footprint.

\paratitle{Layer Sharing} 
In Transformer~\citep{transformer}, both the encoder and decoder are stacks of Transformer layers. Thus, a simple and straightforward way to share the weights in a Transformer is to share them across all Transformer layers. \citet{dabre2019recurrent} share the weights across all Transformer layers for NMT with minimal performance drop. Universal Transformer~\citep{dehghani2019universal} shares the weights across all layers, allowing for recurrent computation with a dynamic halting mechanism and achieves better performance than the vanilla Transformer. ALBERT~\citep{albert} introduces the idea into pretrained language models for natural language understanding (NLU). Although it cannot reduce the computational overheads and has an inevitable negative effect on performance, this design saves up to 95\% of disk space for storing the model, which can be critical for deployment on mobile devices with limited storage. \citet{takase2021lessons} systematically study strategies for sharing weights across layers. Instead of using the weights of one Transformer layer for all layers, they aim to explore the best way to use the parameters of $M$ layers for $N$ layers ($M < N$). 
\citet{reid2021subformer} introduce a strategy named ``sandwich-style'' parameter sharing, which shares the weights for central layers while leaving the first and last layers independent.

\subsection{Low-Rank Factorization}
The weight matrices in a neural network are often low-rank, indicating redundancy in model weights~\citep{sainath2013low}. Thus, a natural idea is to factorize the weight matrices into two or more smaller matrices to save parameters. A common technique for low-rank factorization is singular value decomposition (SVD). For a matrix $A \in \mathbb{R}^{m \times n}$, there exists $A = U \Sigma V^\mathrm{T}$, where $r \leq \min{\{m, n\}}$ is the rank of $A$; $U \in \mathbb{R}^{m\times r}$, $V \in \mathbb{R}^{n\times r}$ are two orthogonal matrices; $\Sigma \in \mathbb{R}^{r\times r}$ is a diagonal matrix with only the non-zero singular values of $A$. Thus, the space complexity can be effectively reduced from $O(mn)$ to $O(mr + rn)$, improving the storage-efficiency of the model. 

\paratitle{Decomposing Linear Layers} Low-rank factorization can be applied to any linear layer. \citet{grachev2017neural} factorize the weights of an LSTM language model. Following that, \citet{winata2019effectiveness} exploit SVD for both the LSTM cell in a language modeling task and a pretrained LSTM language model, ELMo~\citep{elmo}. This is one of the earliest attempts to compress a pretrained language model. \citet{ma2019tensorized} propose a new self-attention module, namely multi-linear attention, as a substitute for the standard multi-head attention module in a Transformer. They use block-term tensor decomposition (BTD, \citealp{btd}) to factorize multi-head attention. Their results demonstrate comparable performance to the vanilla Transformer while being parameter-efficient. \citet{noach2020compressing} propose a two-stage approach to compress a pretrained language model. In the first stage, they decompose each weight matrix in the pretrained language model with SVD. Then, for the second stage, they fine-tune or use knowledge distillation to refine the weights. \citet{chen2021drone} propose data-aware low-rank compression (DRONE) by exploiting the prior knowledge of the data distribution. Instead of minimizing the reconstruction error of the the weight matrix, they minimize the approximation error of the outputs. DRONE achieves better performance than SVD. Besides, as an alternative to SVD, Kronecker decomposition retains the rank of the matrix and has shown improvement compressing BERT and GPT-2~\citep{tahaei2021kroneckerbert,kroneckergpt}.

\paratitle{Decomposing Embedding} ALBERT~\citep{albert} uses factorization for the embedding layer, which has redundant parameters due to its high input and output dimensions.
Since the power of Transformer mainly comes from its contextual learning ability, the parameters in the token embedding layer are not efficient. It intuitively makes sense to reduce them by factorizing the embedding matrix. \citet{reid2021subformer} propose self-attentive factorized embeddings (SAFE) by adding a small self-attention layer on the basis of linear projection to achieve better performance.

\subsection{Pruning}
\label{sec:pruning}

Pruning~\citep{obd} aims to remove unimportant weights from a neural network to achieve storage-, memory-efficiency, and sometimes also computation- and time-efficiency while preserving model performance. There are two key elements in a pruning method: (1) A \textbf{pruning unit} is the atomic unit to be removed from a model; it can be a single weight (unstructured pruning), an attention head or even a Transformer layer (structured pruning). (2) A \textbf{saliency score} is the criterion for making pruning decisions. Based on whether it uses a gradient and which order of gradient it uses, we can categorize pruning methods to zeroth-order (only considering weight magnitude), first-order and second-order approaches. We summarize some representative pruning methods in Table~\ref{tab:prune}.

\paratitle{Unstructured Pruning} Unstructured pruning removes ``unimportant'' connections in a neural network by setting the corresponding weights to 0. After pruning, the weight matrix often becomes sparse. To exploit the characteristics of a sparse matrix to achieve computation- and memory-efficiency, specialized hardware (\eg sparse tensor cores in Nvidia A100~\citep{mishra2021accelerating}) and software (\eg PyTorch Sparse API\footnote{\url{https://pytorch.org/docs/stable/sparse.html}}) are necessary. \citet{see2016} uses magnitude-based pruning with retraining to compress RNN models for NMT. Magnitude-based pruning~\citep{han2016deep} simply prunes weights with smallest magnitude (\ie absolute values). After pruning, \citet{see2016} continue to fine-tune the pruned network to obtain better performance. \citet{narang2017exploring} prune an RNN model gradually during training. The magnitude threshold for pruning is gradually increased with increasing training steps. \citet{wang2020on} first prunes and retrains NMT models with magnitude pruning and then restores the pruned parameters to train the entire network again, in order to achieve better performance than the original model. \citet{zhang2020one} propose a one-shot pruning technique based on the Jacobian spectrum. Different from iterative pruning methods, one-shot pruning techniques only prune a network once and then use standard training to train the sparse network.

Some recent works target transfer learning as it has become the new standard paradigm in NLP. 
\citet{gordon2020compressing} aim to reveal how pruning affects transfer learning. They find that low levels of pruning (30\%--40\%) do not affect pretraining loss or transfer to downstream tasks at all. However, further pruning has a negative 
impact
on both pretraining and transfer learning. A high level of pruning can even prevent the model from fitting downstream datasets.
\citet{sanh2020movement} claim that magnitude pruning is suboptimal for transfer learning. They propose \textit{movement pruning} as a simple first-order method for fine-tuning of pretrained language models. Instead of preserving weights that are currently far from zero, they retain those that are moving away from zero (\ie gaining larger magnitude) during fine-tuning, achieving better performance for pruning BERT.
\citet{guo2021parameter} propose \textit{diff pruning}, by learning a task-specific ``diff'' vector that extends the original pretrained parameters. The task-specific ``diff'' vectors are trained with $L_0$ regularization~\citep{l0} to encourage sparsity. By updating only 0.5$\%$ of parameters, diff pruning achieves similar performance to fine-tuning the whole network. 

\paratitle{Structured Pruning}
Structured pruning removes weight blocks, rows, attention heads, or layers from a model. Compared to unstructured pruning, it can usually achieve acceleration and memory reduction without specialized hardware or software.
\citet{narang2017block} extends gradual pruning for RNNs~\citep{narang2017exploring} to structured pruning. They first divide weights into blocks, then prune blocks of weights in a layer using group lasso regularization~\citep{yuan2006model} to create blocks of zeros in weight matrices. \citet{16heads} and \citet{voita2019analyzing} find that the multi-head attention in Transformer has redundancy. Both works use a first-order approach to remove attention heads from Transformer. Following this, \citet{drophead} propose a structured dropout strategy named LayerDrop. When training a Transformer, a random dropout is applied to each layer. After  one-off training, the model can be pruned on-demand to achieve the target inference speed. 
SNIP~\citep{snip} removes unimportant non-linear terms in the residual connections, whose magnitude is below a threshold.
\citet{lagunas2021block} introduce a block pruning approach that extends structured methods by considering blocks of any size. 
They integrate this structure into movement pruning~\citep{sanh2020movement} and find this approach can automatically learn to prune out full components in Transformer, \eg an attention head. \citet{cofipruning} propose CoFi, a pruning method that jointly prunes both coarse-grained units (\eg layers) and fine-grained units (\eg attention head and hidden units). CoFi also introduces a distillation loss to further improve its performance.

\paratitle{Lottery Ticket Hypothesis} \citet{lottery} propose the ``lottery ticket hypothesis'': dense, randomly-initialized, feed-forward
networks contain subnetworks (winning tickets) that --- when trained in isolation --- reach test accuracy comparable to the original network in a similar number of
iterations. It reveals that retraining from remaining weights~\citep{han2016deep} in a pruned network is not necessary for the pruned network. In contrary, a ``winning ticket'' can always learn better, even when training from scratch (as long as it is initialized with the same random weights). Following this, \citet{chen2020lottery} verify the lottery ticket hypothesis on BERT with iterative magnitude pruning. They find that subnetworks found on the pretraining task (\ie masked language modeling, MLM) transfer universally to downstream tasks whereas those found on downstream tasks do not. \citet{prasanna2020when} also verify the lottery ticket hypothesis with BERT, for both magnitude and structured pruning. They find that even the worst subnetwork in BERT remains highly trainable, suggesting that the weights of BERT may have relatively low redundancy. This seems to be consistent with previous finding on over-parameterized models~\citep{nakkiran2020deep}.

\subsection{Quantization}
Quantization aims to compress a neural network by reducing the number of bits (\ie precision) in the weights of the model, improving storage-, memory-, computation-, and time-efficiency (on most hardware). In general, quantization can be further categorized into post-training quantization (PTQ) and quantization-aware training (QAT). PTQ rescales the weights of a trained model whereas QAT introduces the rounding error into the training process. Due to the considerable performance drop for PTQ, most works in compressing NLP models unanimously use QAT, in order to achieve comparable performance with the full-precision model.

\paratitle{8-Bit Quantization} Quantizing models from full precision floats (FP32) to 8-bit integers (INT8) is a classical setting, since operations including matrix multiplication can be calculated much faster with INT8 than FP32, especially on a CPU.
\citet{zafrir2019q8bert} use symmetric linear quantization~\citep{jacob2018quantization} to quantize both weights and activations to INT8 dynamically. They also explore quantization-aware training (QAT) for BERT. They use \textit{fake quantization}~\citep{jacob2018quantization} to introduce quantization error into the model during training phase to simulate the rounding effect. They use Straight-Through Estimator (STE)~\citep{bengio2013estimating} to estimate the gradient of the non-differentiable fake quantization. They find that dynamic post-training quantization hurts the downstream performance slightly while QAT achieves comparable performance to the original model. Similarly, \citet{prato2020fully} apply QAT with STE for Transformers on neural machine translation and achieve results that are similar to the original model. 
I-BERT~\citep{ibert} eliminates floating point calculation in activation by exploiting lightweight integer-only approximation methods for non-linear operations (\eg GELU, Softmax and LayerNorm) in BERT. The resulting I-BERT model is capable of doing pure INT8 inference thus has a better acceleration ratio.

\begin{figure}[t]
\centering
  \includegraphics[width=\linewidth]{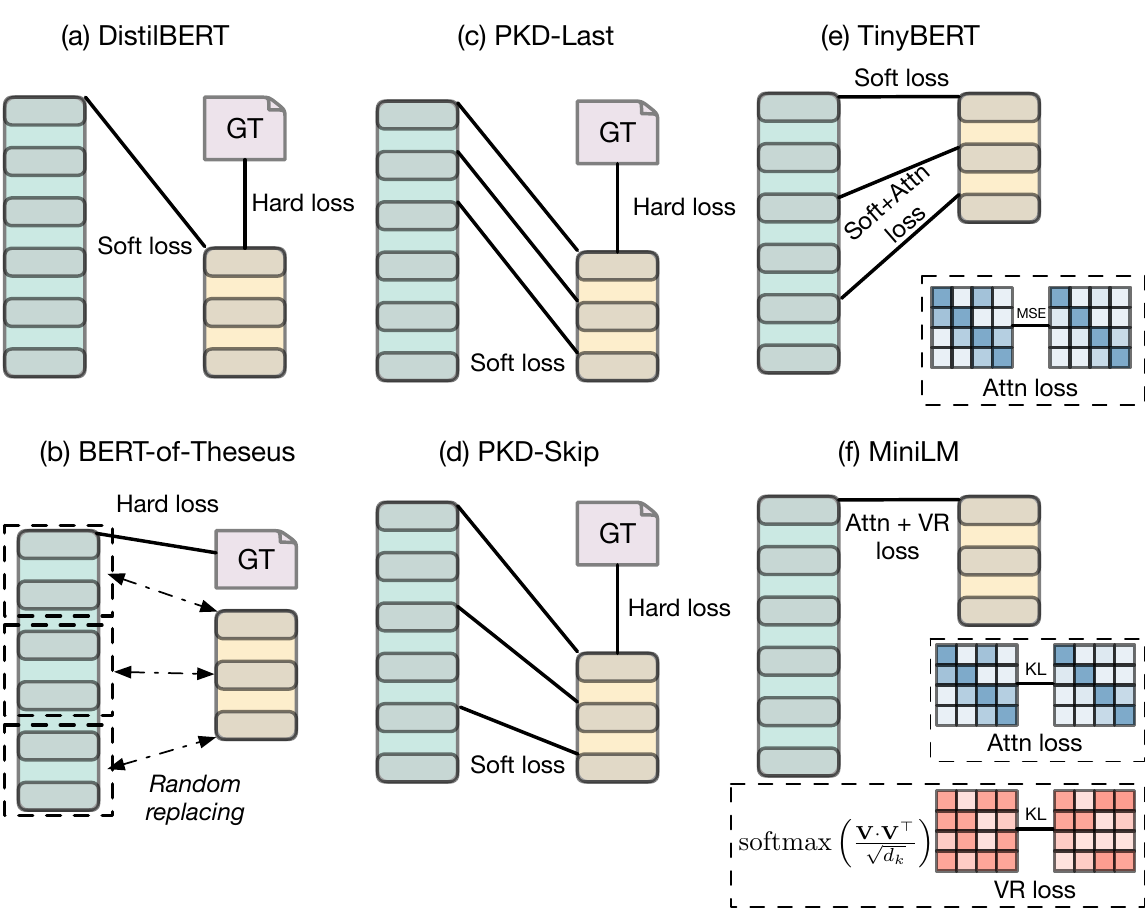}
  \caption{A summary of different KD approaches. ``GT'' represents ground-truth labels.}
  \label{fig:kd}
\end{figure}

\paratitle{Lower-Bit Quantization} Recent works aim to push quantization to even lower precision. Lower-bit quantization faces more challenges, including difficulty to optimize, and lack of model expressivity. 
\citet{qbert} propose a group-wise quantization scheme and use second-order Hessian-based mixed-precision method~\citep{hawq} to quantize BERT down to 2 bits. They claim that weights corresponding to each neuron could lie in different ranges of real numbers. For example, for a multi-head self-attention module, they split the weight matrix to 12 groups, with respect to each attention head. Then they further split each group and have a total number of 128 subgroups, each of which has its own quantization range. GOBO~\citep{gobo} separates the weights into two groups --- \textit{Gaussian} and \textit{outliers} where the former is quantized to 3 bits and the latter remains 
a
full-precision float (FP32). TernaryBERT~\citep{ternarybert} combines approximation-aware and loss-aware ternarization (\ie using only $\{-1,0,1\}$ for weights) methods with different granularity for different components in BERT. They further add knowledge distillation to improve the performance of QAT. \citet{binarybert} observe a large performance drop from a ternary network to a binary network when trained directly, due to its loss landscape. They propose ternary weight splitting, which initializes BinaryBERT by
splitting a half-sized ternary network into two binary networks. The initialized network inherits good performance and can be further fine-tuned without optimization difficulties. \citet{tao2022compression} analyze the reasons why quantization is less effective on generative LMs (e.g., GPT-2, BART). They conclude that homogeneous word embeddings caused by reduced capacity and varied distribution of weights are responsible for the failure. They propose a token-level contrastive distillation and a module-wise dynamic scaling mechanism to mitigate these two problems.

\subsection{Knowledge Distillation}
Knowledge Distillation~\citep{hinton2015distilling} is a widely used technique to transfer knowledge from a large model (teacher) to a smaller one (student) to achieve all types of efficiency. KD usually requires designing a loss function to minimize the distance of the output or intermediate features between the student and the teacher. As illustrated in Figure~\ref{fig:kd}, we summarize the designs of loss functions used in recent works distilling NLP models. Based on the loss function designs, we can further categorize the methods into logit-based KD, feature-based KD, KD with a dynamic target, and module replacing.

\paratitle{Logit-based KD} Following \citet{hinton2015distilling}, logit-based KD methods are the first attempts to distill a large pretrained language model into a smaller one to improve its efficiency. Logit-based KD uses the KL divergence or mean squared error (MSE) to minimize the logits between the student and the teacher. \citet{tang2019distilling} distills fine-tuned BERT into a BiLSTM model in a task-specific setting. The resulting BiLSTM outperforms its counterpart trained without KD by a large margin. DistilBERT~\citep{sanh2019distilbert} distills BERT in a pretraining setting on the task of masked language modeling (MLM). The loss is a combination of three components: the original MLM loss, cosine distance, and KL divergence. After distillation, the model can be fine-tuned and perform downstream tasks. \citet{turc2019well} explore the effect of initialization for the student. They find that a student BERT pretrained with MLM outperforms random initialization and truncated teacher~\citep{sanh2019distilbert,pkd} when used to initialize the student model. \citet{mixkd} use MixUp~\citep{mixup} for data augmentation to distill BERT.

\paratitle{Feature-based KD}
Instead of using only the final output, feature-based KD aims to align the intermediate features between the teacher and the student. PKD~\citep{pkd} introduces an MSE loss between layer representations. As shown in Figure~\ref{fig:kd}(c) and (d), they propose two strategies: one aligns the student with the last few layers in the teacher (PKD-Last) and the other learns the teacher's representations in every 2 layers (PKD-Skip). The latter strategy performs slightly better in experiments. A similar technique is also presented in \citet{aguilar2020knowledge}.
On the basis of PKD, TinyBERT~\citep{tinybert} further introduces an attention loss that aims to align the attention matrices in layers between the teacher and the student, as illustrated in Figure~\ref{fig:kd}(e). TinyBERT also demonstrates that performing KD in both pretraining and fine-tuning stages can improve the performance of KD.
Similarly, MiniLM~\citep{minilm,minilmv2} aligns the attention matrix and value-value scaled dot-product (\ie value relation loss, as shown in Figure~\ref{fig:kd}(f)). The added feature complements the attention matrix (\ie queries-keys scaled dot-product) and allows the complete transfer of multi-head self-attention.
\citet{wu2021one} propose a multi-teacher distillation framework that use both intermediate features and soft labels from multiple teachers to distill a student and achieve better performance.
DynaBERT~\citep{dynabert} uses layer-wise KD loss to distill a teacher into a student model that has sub-networks of different widths and depths. Thus, the same model can be used on various devices with different computing budgets.
MobileBERT~\citep{mobilebert} redesigns a BERT architecture that is suitable for mobile devices. In addition to the layer-wise feature distillation~\citep{pkd} and attention distillation~\citep{tinybert}, they introduce a progressive knowledge transfer mechanism by distilling the model layer by layer, instead of altogether. \citet{liu2022multi} exploit structural features on both token and span levels to align the student with the teacher.

\paratitle{Module Replacing} A special case of KD is BERT-of-Theseus~\citep{bot}. As shown in Figure~\ref{fig:kd}(b), BERT-of-Theseus does not apply any knowledge transfer loss to minimize the distance between the student and the teacher. Instead, they freeze the teacher modules and train a hybrid model by randomly replacing some modules in the teacher model with student modules. They further design a linear scheduler to increase the probability of replacement to bridge the gap between training and inference. Following this, Sparse Progressive Distillation~\citep{huang2022sparse} uses layer-wise KD to iteratively prune the student modules while randomly replacing each module in the teacher model with its corresponding student module with a fixed probability. After the target sparsity is hit, the replacing rate is progressively increased to 1. This method combines feature-based KD, module replacing, and pruning, achieving a super-teacher performance on GLUE~\citep{glue}.

\paratitle{KD with Dynamic Targets} In traditional KD, the teacher serves as a static target for the student to match, without any update during distillation. However, this can be suboptimal since the teacher is unaware of the student or its goal to transfer the knowledge to the student. ProKT~\citep{shi2020learning} projects the supervision signals of a teacher model into the student's parameter space by decomposing the training objective into local intermediate targets with approximate mirror descent~\citep{beck2003mirror}. \citet{zhou2021meta} propose a simpler framework with meta learning to allow the teacher to adapt itself for better knowledge transfer. The student is evaluated on a ``quiz'' set after a few training steps and provides feedback to the teacher. Its first-order variant can further improve training speed and reduces memory footprint~\citep{raptilemeta}.

\begin{table}[t]
\centering
\resizebox{1.\linewidth}{!}{
\begin{tabular}{lll}
\toprule
\textbf{Method} & \textbf{Exit criterion} \\
\midrule
DeeBERT~\shortcite{deebert} & entropy $<\theta$ \\
RightTool~\shortcite{schwartz2020right} & calibrated max class probability $>\theta$ \\
FastBERT~\shortcite{fastbert} & entropy $<\theta$   \\
RomeBERT~\shortcite{geng2021romebert} & entropy $<\theta$   \\
SkipBERT~\shortcite{skipbert} & max class probability $>\theta$ \\
\midrule
PABEE~\shortcite{pabee} & patience (\#consistent prediction $>\theta$) \\
Voting~\shortcite{sun2021early} & accumulated votes $>\theta$  \\
LeeBERT~\shortcite{leebert} & patience (\#consistent prediction $>\theta$) \\
Past-Future~\shortcite{liao2021global} & entropy $<\theta$  \\
PCEE-BERT~\shortcite{pceebert} & patience (\#consistent IC confidence $>\theta$) \\
\midrule
BERxiT~\shortcite{xin2021berxit} & estimated confidence $>\theta$  \\
CAT~\shortcite{schuster2021consistent} & estimated conformity $>\theta$  \\
\bottomrule
\end{tabular}
}
\caption{A summary of three types of early exit methods: confidence estimation, internal ensemble, and learning to exit. $\theta$ is a predefined threshold for exiting.}
\label{tab:ee}
\vspace{-1em}
\end{table}

\subsection{Early Exit}
\label{sec:ee}
Early exit (EE) does not reduce the size (the total number of parameters) of the model. Instead, EE accelerates model inference by terminating inference at a particular layer based on some criteria. Although it does not make a model smaller it can reduce computation and achieve acceleration.
Early exit inserts internal classifiers (which are often simple linear layers) into a large network as triggers for early exiting. The key element in early exit methods is the exit criterion. There are three types of exit criteria: confidence estimation, internal ensemble and learning to exit. We summarize the exit criteria in Table~\ref{tab:ee}.

\paratitle{Confidence Estimation} Previous works in computer vision~\citep{park2015big,branchynet,shallowdeep} define a metric as the proxy for confidence of prediction. The inference can exit early if the confidence reaches a threshold at an early layer. This idea is then applied to pretrained LMs~\citep{deebert}. For each Transformer layer, a linear internal classifier (IC) is inserted after the Transformer layer. When doing inference, the model exits early when an IC outputs a predicted probability with an entropy lower than the threshold. A similar approach is proposed in RightTool~\citep{schwartz2020right}. The temperature-calibrated maximum class probability is used as confidence. FastBERT~\citep{fastbert} distills the output final classifier into earlier classifiers for better performance. Following that, RomeBERT~\citep{geng2021romebert} proposes gradient regularization to facilitate the KD. SkipBERT~\citep{skipbert} replaces lower BERT layers with pre-computed representation of text chunks and uses confidence-based early exit for higher layers to achieve maximum acceleration.

\paratitle{Internal Ensemble}
A weakness of confidence estimation is poor utilization of computation. When confidence of an internal classifier fails to satisfy the exit criterion, all relevant computation becomes invalid. Reusing the results from previous layers to improve the qualify of early exit can be a promising direction. Internal ensemble approaches consider outputs and predictions from multiple internal classifiers to make better decisions. This is similar to ensemble learning, only it is within a single model.

The first work of internal ensemble, PABEE~\citep{pabee}, draws a comparison between overfitting in training and overthinking in inference and adapts early stopping for inference. When doing inference, the model will exit once multiple consecutive internal classifiers make the same prediction. The threshold, namely patience, is a hyperparameter that can be adjusted to achieve different trade-off between accuracy and speed. Besides improvement on performance and efficiency, PABEE achieves better adversarial robustness, which the authors attribute to the ensemble effect of internal ensemble. \citet{sun2021early} propose a diversity loss that encourages ICs to have diverse probability distributions. Then, they use a voting mechanism to internally ensemble the classifiers. Every IC has one vote in final prediction. The model will exit when one class has accumulates enough votes. LeeBERT~\citep{leebert} promotes consistency of IC predictions by distilling them mutually. Then, it follows PABEE's patience-based strategy to decide when to exit. \citet{liao2021global} introduce a more elaborate mechanism for internal ensemble. They first train ``imitation learners'', which are linear layers that predict the hidden states of future layers based on hidden states that are already calculated. PCEE-BERT~\citep{pceebert} combines patience-based exit with confidence estimation and terminates inference when multiple layers are confident.

\paratitle{Learning to Exit} 
Other works use a learning-based approach to make exit decisions. BERxiT~\citep{xin2021berxit} trains a linear learning-to-exit (LTE) module to predict whether the current IC prediction is correct. CAT~\citep{schuster2021consistent} proposes a ``meta consistency classifier'' to predict whether the current IC prediction matches the final classifier and exits when the consistency classifier predicts a level of conformity higher than the threshold.

\begin{table}[t]
    \vspace*{0.1cm}
    \centering
    \resizebox{\linewidth}{!}{
    \begin{tabular}{llll}
    \toprule
      \textbf{Method}         & \textbf{Optimization}   & \textbf{Feature}     & \textbf{Skipped Tokens}                         \\ \midrule
    
    PoWER~\shortcite{powerbert}    & soft masking                 &  attention & discarded               \\
    TR-BERT~\shortcite{trbert}     & RL        &         hidden states & forwarded\\
    
    LAT~\shortcite{lat} & soft masking  & attention & forwarded           \\
    
    LTP~\shortcite{ltp}     & soft masking     & attention & discarded         \\
    
    Transkimmer       & reparameterization            & hidden states & forwarded              \\ \bottomrule
    \end{tabular}
    }
    \caption{A summary of token skipping methods. This table is adapted from \citet{transkimmer}.}
    \label{tab:token_skipping}
\end{table}

\subsection{Token Skipping}
Similar to early exit, token skipping dynamically accelerates a PLM without reducing its size. The general idea is to skip some tokens for higher layers based on their importance. A summary of these methods is shown in Table~\ref{tab:token_skipping}.

PoWER-BERT~\citep{powerbert} drops a portion of tokens between each Transformer layer based on the attention received by each token. The number of tokens to drop at each layer (\ie \textit{schedule}) is learned by jointly optimizing the sparsity of a soft mask layer with the original loss function. This approach obtains better Pareto curves of accuracy-time trade-off. TR-BERT~\citep{trbert} introduces a dynamic mechanism for making decisions of skipping tokens. It is trained with reinforcement learning with a reward that promotes classifier confidence and penalizes the number of retained tokens. Different from PoWER-BERT, the skipped tokens are forwarded to the last layer instead of getting removed. Length-Adaptive Transformer (LAT, \citealp{lat}) introduces LengthDrop that randomly skips tokens during pretraining to mitigate the gap between pretraining and fine-tuning. The schedule of LAT is searched with an evolutionary search algorithm. LTP~\citep{ltp} learns a threshold for each Transformer layer. Instead of following a schedule to drop a specific number of tokens, LTP simply drops tokens with a saliency score (received attention) lower than the learned threshold. Transkimmer~\citep{transkimmer} adds a skim predictor module, consisting of a small MLP and Gumbel-Softmax reparameterization before each layer. The skim predictors output a mask deciding whether to drop a token. It also employs a skim loss that optimizes the ratio of skipped tokens to the total number of tokens to encourage sparsity.

\section{Challenges and Future Directions}
\begin{figure}[t]
\centering
  \includegraphics[width=\linewidth]{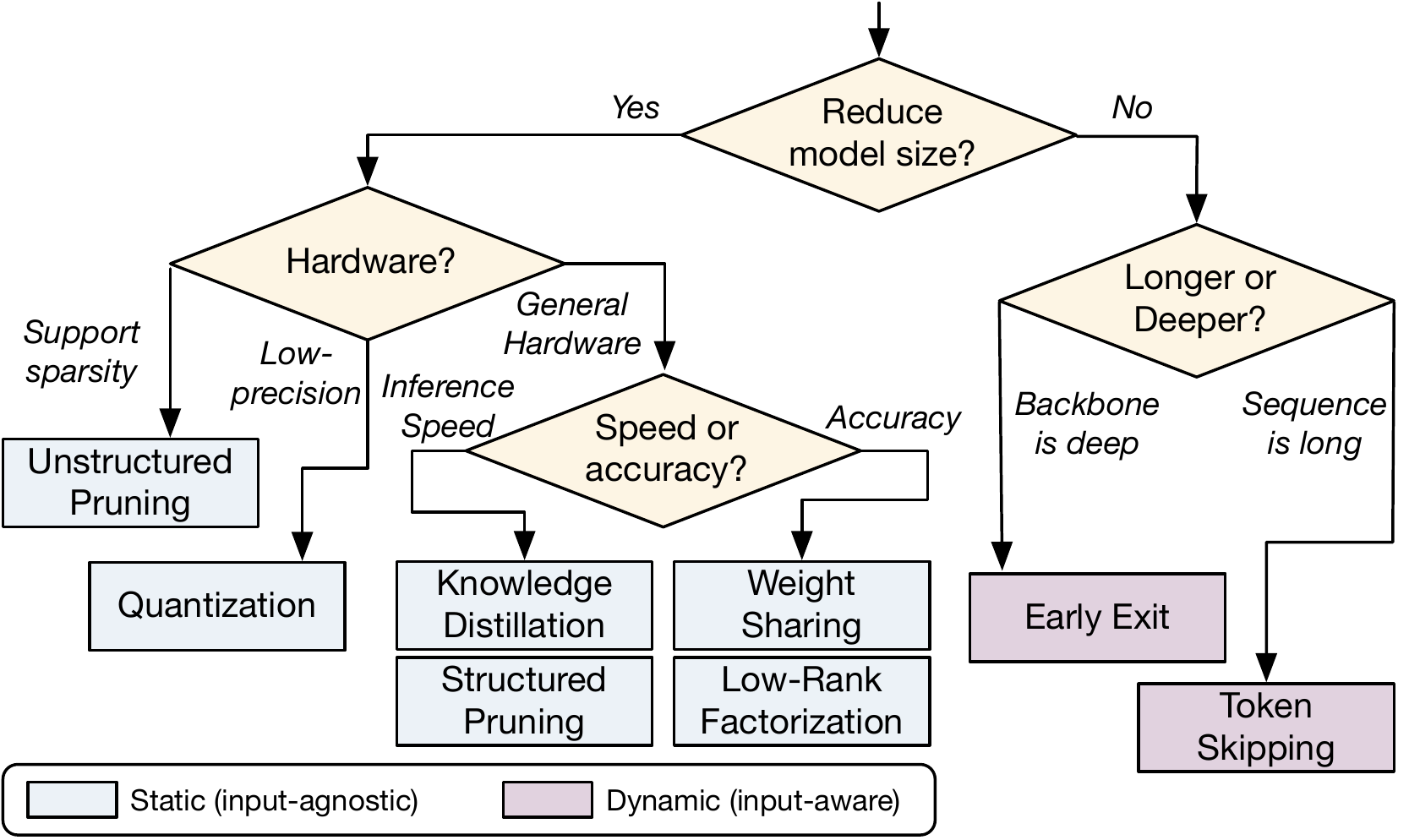}
  \caption{An \textit{oversimplified} decision tree for model compression and acceleration.}
  \label{fig:decision}
\end{figure}

\paratitle{Which Technique to Use?} A common question asked is how to decide which technique to use in practice? Unfortunately, there is no silver bullet given  that we need to take the task, data, backbone, 
and
hardware 
into consideration. We provide an oversimplified decision tree (as shown in Figure~\ref{fig:decision}) only as a starting point.
Note that these techniques can often be combined for better results (to be discussed shortly).

\paratitle{Evaluation} Although there have been benchmarks proposed for evaluating model compression and acceleration as introduced in Section~\ref{sec:benchmarks}, there are several drawbacks in current evaluation. First, there is no generally recognized setting for evaluation of model compression and acceleration. Different studies often yield models with different speed-up ratio, number of parameters and accuracy. Thus, it is often difficult to directly compare them, not to mention differences in hardware. Second, general NLU benchmarks like GLUE~\citep{glue} or SuperGLUE~\citep{superglue} may not be the best to represent more common tasks on a mobile device. Tasks like intention detection, dense retrieval, and spam classification could be more representative.

\paratitle{Combining Techniques} Although there have been attempts at combining multiple model compression and acceleration techniques~\citep{kim2020fastformers,sanh2020movement,xu2021beyond}, there is a lack of comprehensive and systematic study for combining compression techniques for better performance and efficiency. Constructing a best practice to compress a large model can be useful for practitioners.

\paratitle{Explainability and Robustness} Recent works~\citep{stanton2021does,xu2021beyond} cast doubt on the explainability of model compression and acceleration. Meanwhile, recent works~\citep{du2021compressed,xu2021beyond} report negative effects of model compression on robustness. Explainable and robust compression methods can be important for applications of model compression and acceleration. Also, explainable and robust compression minimizes effort to re-evaluate the compressed model, and thus can be reliable and predictable in production~\citep{stanton2021does,xu2021beyond}.

\paratitle{Minimizing Human Effort} Current compression and acceleration approaches still largely rely on human heuristics to achieve good performance. For example, knowledge distillation often requires an elaborately designed loss function; pruning relies on the saliency score; weight sharing and low-rank factorization involve expertise to appoint modules for sharing or factorization. One promising direction could be applying Meta Learning~\citep{maml} or Neural Architecture Search~\citep{darts} to model compression and acceleration, to minimize the need for hyperparameters and human design. 

\section*{Acknowledgments}
We appreciate the insightful comments from the anonymous reviewers. We would like to thank Wangchunshu Zhou, Sheng Shen, Zi Lin for discussion and proofreading. This work is supported by NSF Award \#1750063.

\bibliography{custom.bib}

\end{document}